\documentclass{article} 
\PassOptionsToPackage{table}{xcolor}
\usepackage[final]{colm2026_conference}

\usepackage{microtype}
\usepackage{hyperref}
\usepackage{url}
\usepackage{booktabs}
\usepackage{amsmath}
\usepackage{amssymb}
\usepackage{graphicx}
\usepackage{cleveref}

\usepackage{lineno}

\definecolor{darkblue}{rgb}{0, 0, 0.5}
\hypersetup{colorlinks=true, citecolor=darkblue, linkcolor=darkblue, urlcolor=darkblue}

\usepackage{booktabs}   
\usepackage{multirow}
\usepackage{makecell}       
\usepackage{diagbox}        
\usepackage{xcolor} 
\definecolor{lightgray}{gray}{0.85}
\definecolor{lightblue}{RGB}{240, 245, 255}

\title{Multi-objective Evolutionary Merging \\ Enables Efficient Reasoning Models}

\author{Mario Iacobelli~$^{1*}$ \And Adrian R. Minut~$^{2**}$ \And
Tommaso Mencattini~$^3$ \And Donato Crisostomi~$^2$ \And Andrea Santilli~$^4$\thanks{ Work done while at Sapienza University of Rome. $^{**}$~Correspondence to: \texttt{minut@di.uniroma1.it}\\ \begin{center}\vspace{-0.17in}
    $^1$~Independent Researcher \, $^2$~Sapienza University of Rome \, $^3$~EPFL \, $^4$~NVIDIA \, $^5$~Paradigma
\end{center}} \And Iacopo Masi~$^2$ \And Emanuele Rodolà~$^{2,5}$
}

\begin{document}

\ifcolmsubmission
\linenumbers
\fi

\maketitle

\begin{abstract}
Reasoning models achieve strong performance on complex problems by leveraging long chains of thought, but this deliberate reasoning incurs substantial inference-time cost. The Long-to-Short (L2S) reasoning problem seeks to preserve accuracy while reducing generated tokens. Yet, current training-free model merging approaches rely on brittle, fixed-hyperparameter arithmetic methods that force suboptimal compromises. We introduce \textbf{Evo-L2S}, a multi-objective evolutionary model merging framework that explicitly optimizes accuracy and output length to recover a Pareto front of merged models. To make this search computationally tractable, we propose an entropy-based subset sampling technique that substantially reduces fitness-estimation overhead. Across six mathematical reasoning benchmarks, Evo-L2S reduces reasoning length by over 50\% at the 1.5B and 7B scales while preserving or improving problem-solving accuracy; at 14B, the steeper Pareto front reveals that attainable compression depends on the intrinsic compressibility of the reasoning model. Overall, Evo-L2S shows that reasoning models can be made substantially more concise while preserving strong problem-solving performance.

\end{abstract}

\begin{figure}[ht]
  \centering
  \includegraphics[width=0.73 \linewidth]{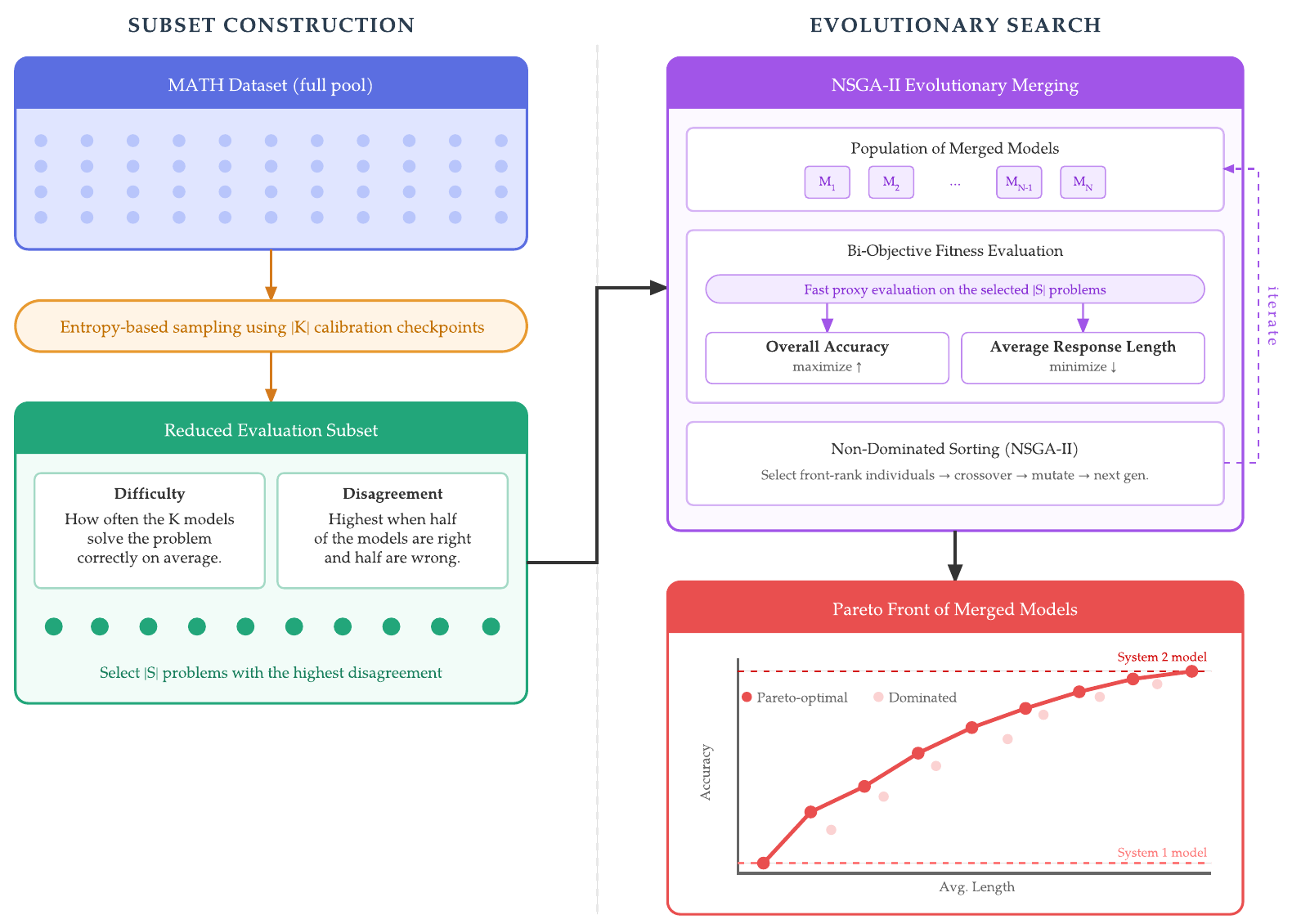}
  
  \caption{Overview of the Evo-L2S pipeline. Left: Entropy-based sampling constructs an informative evaluation subset. Right: Multi-objective evolutionary search (NSGA-II) optimizes reasoning accuracy and output length to produce a Pareto front of merged models.}
  \label{fig:pipeline}
\end{figure}

\section{Introduction}

Recent progress in Large Language Models (LLMs) has led to a paradigm shift from efficient, implicit reasoning toward more deliberate and structured reasoning. This transition is often interpreted through the lens of dual-process theory, a psychological framework that defines human cognition as the interplay between fast, automatic, intuitive thinking (System 1) and slower, more analytical, and deliberate thinking (System 2) \citep{li202512surveyreasoning}. 

While System 2 reasoning has significantly improved performance on complex mathematical and logical tasks, it introduces a substantial computational overhead. The generation of long chain-of-thought (CoT) traces often results in redundant intermediate steps, repeated hypothesis exploration, and unnecessary deliberation. This phenomenon, commonly referred to as overthinking \citep{chen2025think23overthinkingo1like}, leads to increased inference latency and computational cost without proportional accuracy gains on problems that do not require deep reasoning. As LLM deployment scales, this inefficiency becomes a critical bottleneck.

This trade-off between reasoning robustness and inference efficiency is at the core of the Long-to-Short reasoning (L2S) problem \citep{wu2025unlockingefficientlongtoshortllm}: retaining the accuracy benefits of System 2 reasoning while substantially reducing the length of CoT traces and the associated computational cost. Previous training-free attempts to address L2S via model merging rely on scalarized, fixed-hyperparameter arithmetic methods (e.g., Task Arithmetic, TIES) or require highly sensitive initialization (e.g., ACM). These approaches are fundamentally brittle: they force a premature compromise between competing objectives and require exhaustive manual tuning, often collapsing into suboptimal trade-offs.

To overcome this, we introduce \textbf{Evo-L2S}, a novel framework that formulates Long-to-Short reasoning as a multi-objective optimization problem. By leveraging evolutionary model merging, Evo-L2S autonomously explores the parameter space to approximate the Pareto frontier between reasoning accuracy and output length.

Specifically, we make the following contributions:
\begin{itemize}
    \item \textbf{Multi-Objective Formulation for L2S:} We introduce Evo-L2S, a training-free merging procedure that explicitly optimizes the trade-off between accuracy and output length. By combining System 2 and System 1 models, we generate a Pareto-optimal family of merged models, eliminating the need for brittle hyperparameter guessing.
    \item \textbf{Scalable Entropy-Based Fitness Estimation:} To make evolutionary search computationally tractable over massive LLMs, we propose a theoretically grounded, entropy-based subset sampling technique. This drastically reduces the computational overhead of fitness estimation by identifying the most informative evaluation items, ensuring high ranking fidelity at a fraction of the cost.
    \item \textbf{Empirical Validation at Scale:} Using our standardized, \texttt{QwenLM}-based evaluation pipeline to jointly assess accuracy and token efficiency, we conduct comprehensive experiments across the 1.5B, 7B, and 14B parameter scales on six rigorous mathematical benchmarks. At 1.5B and 7B, Evo-L2S reduces reasoning length by over 50\% while preserving strong problem-solving performance; at 14B, we observe a steeper accuracy–compression trade-off.
\end{itemize}

\begin{figure}[ht]
  \centering
  \includegraphics[width=\linewidth]{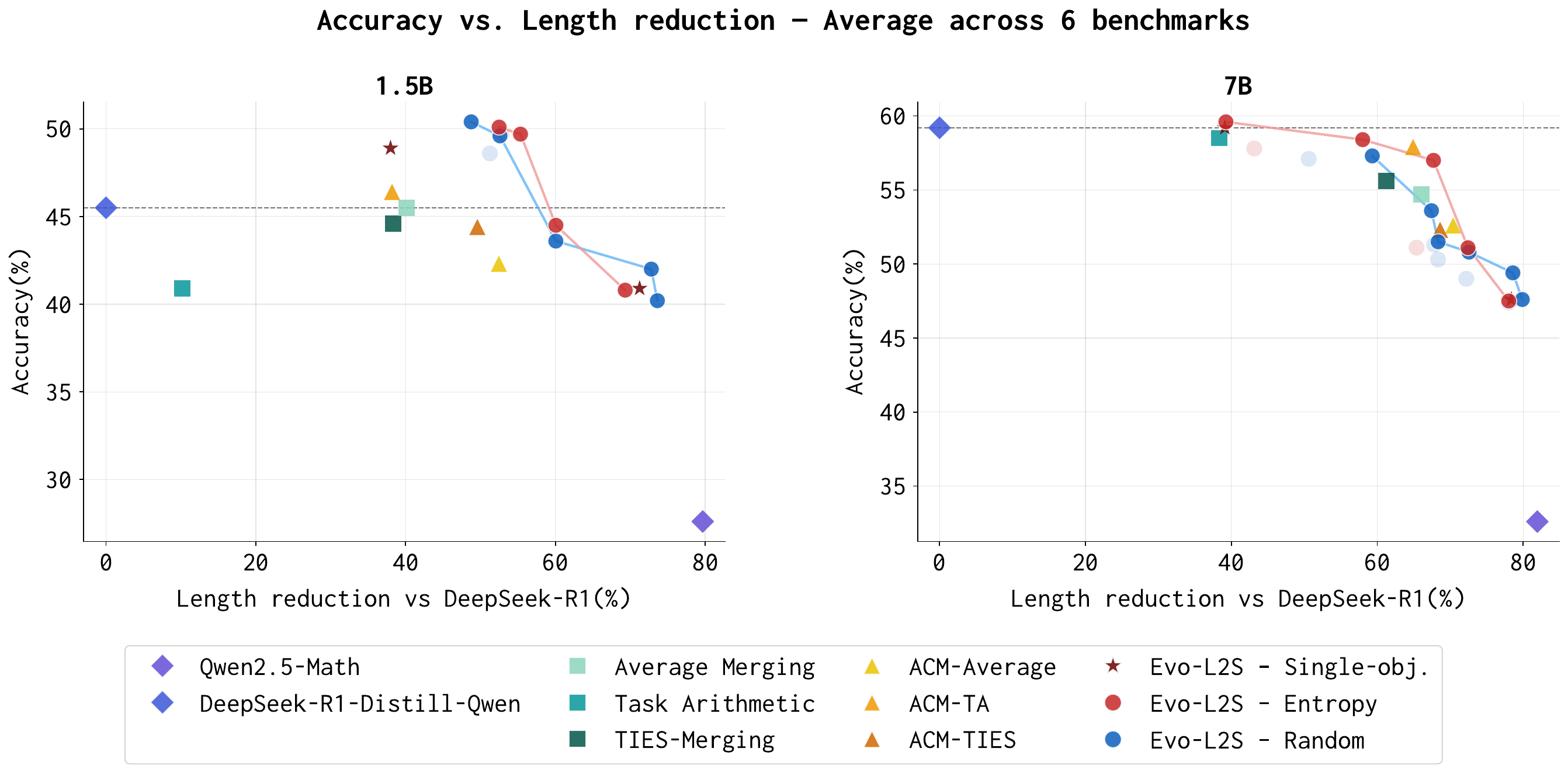}
  \caption{Accuracy (\%) vs.\ length reduction (\%) relative to \texttt{DeepSeek-R1-Distill-Qwen}
    at the \textbf{1.5B} (left) and \textbf{7B} (right) scales, averaged across six reasoning benchmarks. The dashed line marks the \texttt{DeepSeek-R1} accuracy baseline. \textbf{Connected points form the Pareto fronts found by Evo-L2S}; transparent points are Pareto-optimal on the fitness subset but dominated on the full evaluation. The optimal region is in the upper right corner.}
  \label{fig:pareto}
\end{figure}

\section{Related Work}

\subsection{L2S: Long-to-Short reasoning}

Improving the efficiency of System~2 reasoning has motivated approaches operating at different stages of the model lifecycle.

\paragraph{Training-based methods.}
SFT-based approaches directly learn shorter reasoning traces. TokenSkip~\citep{xia2025tokenskipcontrollablechainofthoughtcompression} learns to skip less informative CoT tokens, C3oT~\citep{kang2024c3otgeneratingshorterchainofthought} trains jointly on compressed and original traces, and CoT-Valve~\citep{ma2025cotvalvelengthcompressiblechainofthoughttuning} learns a parameter-space direction controlling CoT length. RL-based methods instead incorporate explicit length incentives: O1-Pruner~\citep{luo2025o1prunerlengthharmonizingfinetuningo1like}, DAST~\citep{shen2025dastdifficultyadaptiveslowthinkinglarge}, and ThinkPrune~\citep{hou2025thinkprunepruninglongchainofthought} use reward shaping or adaptive token budgets to encourage concise reasoning while preserving accuracy.

\paragraph{Inference-time methods.}
Prompt-based approaches such as Concise CoT~\citep{nayab2025concisethoughtsimpactoutput} reduce verbosity through prompting, while ThinkLess~\citep{li2025thinkless} and DEER~\citep{yang2025deer} control reasoning dynamically during decoding, including through early-exit mechanisms. These methods reduce reasoning cost without modifying model weights, but may introduce additional monitoring or stopping decisions for inference-time intervention.

\subsection{Model Merging}
Simple parameter averaging~\citep{wortsman2022modelsoups} can smooth noise when models are well aligned but degrades when checkpoints diverge. Task Arithmetic~(TA;~\citealp{ilharco2023editingmodels,ortiz2023task,zhou2025atmimprovingmodelmerging}) scales and combines task vectors -- the parameter displacements induced by fine-tuning -- via a single coefficient $\lambda$. TIES-Merging~\citep{DBLP:conf/nips/YadavTCRB23} adds conflict resolution through magnitude trimming and sign election, while DARE~\citep{yu2024dare} sparsifies task vectors via random dropping and rescaling. However, all arithmetic methods share a key limitation: they rely on globally fixed hyperparameters, and even small variations can cause significant changes in accuracy and output length, while the same fixed settings lead to markedly different results across model scales. Activation-guided Consensus Merging (ACM;~\citealp{yao2025activationguided}) addresses the uniformity issue by computing layer-wise coefficients from the mutual information between pre-trained and fine-tuned activations on a calibration corpus. Yet ACM is not a stand-alone procedure: it operates as a plug-and-play refinement on top of a pre-merged checkpoint produced by an arithmetic method, and from our experiments, the final accuracy--length trade-off is highly sensitive to this initialization. Under unfavorable pre-merge choices, ACM can even degrade performance.

The inherent sensitivity of these arithmetic methods is particularly pronounced in the Long-to-Short (L2S) reasoning paradigm, where the objective is to collapse expansive Chain-of-Thought (CoT) trajectories into concise, logically sound outputs without sacrificing symbolic accuracy. \citet{wu2025unlockingefficientlongtoshortllm} first established model merging as a viable, training-free alternative for L2S, demonstrating that weight-space interpolation can effectively mitigate the redundant "overthinking" reflections common in large reasoning models. While \citet{yao2025activationguided} subsequently applied ACM to L2S to provide more granular control, the dependence on a stable arithmetic initialization remains a bottleneck. This highlights a critical research gap: the need for an optimization strategy capable of autonomously navigating the non-linear Pareto frontier between reasoning depth and token efficiency, a high-dimensional search problem that standard heuristics are ill-equipped to solve.

\section{Preliminaries}
\label{sec:preliminaries}
\subsection{Formulating L2S as Multi-Objective Optimization}

We formulate Long-to-Short (L2S) reasoning as a multi-objective optimization problem. Instead of searching for a single merged model that balances accuracy and brevity under a fixed set of hyperparameters, we jointly optimize two conflicting objectives: (i) \textbf{accuracy} (Pass@1), defined as the fraction of problems solved correctly in a single attempt (which, under greedy decoding, coincides with standard accuracy), and (ii) \textbf{output length}, measured as the mean number of tokens generated per response and used as a proxy for inference-time computational cost.

To this end, we adopt Mergenetic~\citep{minut-etal-2025-mergenetic}, a library for evolutionary model merging built on top of MergeKit~\citep{goddard-etal-2024-arcees} for parameter-space merging and PyMoo~\citep{pymoo} for evolutionary optimization. A core contribution of this work is extending Mergenetic to support the L2S setting. We implement a custom evaluation pipeline, built on the QwenLM toolkit\footnote{https://github.com/QwenLM/Qwen2.5-Math}, which jointly measures accuracy and response length on mathematical reasoning benchmarks. Additionally, we substitute the default MERGE$^3$~\citep{mencattini2025merge3} with our entropy-based sampling procedure for the reduced evaluation set, avoiding the need to train a \texttt{p-IRT} model.

\subsection{Slow and Fast Thinking Models}
\label{sec:model-pairs}

Our experiments merge pairs of architecturally compatible checkpoints
representing the two reasoning systems, which lie at opposite ends of the
accuracy--length trade-off (\Cref{tab:models}).

\begin{table}[t]
\centering
\small
\begin{tabular}{cll}
\toprule
\textbf{Scale} & \textbf{System~1} & \textbf{System~2} \\
\midrule
1.5B & \texttt{Qwen2.5-Math-1.5B} & \texttt{DeepSeek-R1-Distill-Qwen-1.5B} \\
\texttt{7B}   & \texttt{Qwen2.5-Math-7B}   & \texttt{DeepSeek-R1-Distill-Qwen-7B}   \\
\texttt{14B}  & \texttt{Qwen2.5-14B}       & \texttt{DeepSeek-R1-Distill-Qwen-14B}  \\
\bottomrule
\end{tabular}
\caption{Model pairs used in our experiments, sharing Qwen architectures.}
\label{tab:models}
\end{table}

\paragraph{System~1 endpoint.}
As the fast-thinking endpoint we use the \textbf{Qwen2.5}
family~\citep{qwen2025qwen25technicalreport}.
At the 1.5B and 7B scales we employ the math-specialised variants
\textbf{\texttt{Qwen2.5-Math}} ~\citep{yang2024qwen25mathtechnicalreportmathematical}, obtained by continuing pre-training on the Qwen Math
Corpus~v2 (over one trillion tokens).
These models generate short, concise responses but exhibit lower accuracy on
multi-step reasoning benchmarks.

\paragraph{System~2 endpoint.}
As the slow-thinking endpoint we use the \textbf{\texttt{DeepSeek-R1-Distill-Qwen}}
family.
DeepSeek-R1~\citep{dpsk-r1} is a 671B-parameter Mixture-of-Experts reasoning model trained
through a multi-stage pipeline combining supervised fine-tuning, reinforcement
learning, rejection sampling, and alignment.
Dense distilled variants are obtained by fine-tuning Qwen2.5 backbones on
approximately 800k high-quality CoT traces generated by the full model.
The Qwen-based distilled checkpoints at 1.5B and 7B produce long reasoning traces and achieve high accuracy on complex tasks, at substantial token cost.

\subsection{Merging Operator}
\label{sec:operators}
 
As in the canonical model-merging formulation, our checkpoint pairs derive from a common pre-trained initialization, ~$\theta_0$. This shared ancestry guarantees architectural compatibility between the two endpoints, allowing us to define a single displacement vector:
\begin{equation}
  \tau = \theta_{\mathrm{S1}} - \theta_{\mathrm{S2}},
  \label{eq:tau}
\end{equation}
where $\theta_{\mathrm{S2}}$ denotes the \texttt{DeepSeek-R1-Distill-Qwen} checkpoint
(System~2) and $\theta_{\mathrm{S1}}$ denotes the \texttt{Qwen2.5-Math} checkpoint
(System~1).
Here $\tau$ does not represent a fine-tuning update; it encodes a direct
parameter-space displacement from the slow-thinking model toward the
fast-thinking one.
 
\paragraph{Task Arithmetic (TA).}
Task Arithmetic~\citep{ilharco2023editingmodels} applies the displacement
$\tau$ with a scalar coefficient~$\lambda$, yielding a global linear
interpolation between the two endpoints:
\begin{equation}
  \theta_M = (1 - \lambda)\,\theta_{\mathrm{S2}} + \lambda\,\theta_{\mathrm{S1}},
  \label{eq:ta}
\end{equation}
where $\lambda \in [0,1]$.
When $\lambda = 0$ the merged model coincides with the System~2 endpoint; when
$\lambda = 1$ with the System~1 endpoint.
In our evolutionary framework, $\lambda$ constitutes the \emph{genotype}: the
single decision variable that fully specifies a candidate merge.
 
\paragraph{Alternative merging operators.}
We additionally experiment with:
(i)~\textbf{TIES-Merging}~\citep{DBLP:conf/nips/YadavTCRB23}, which introduces a
density parameter~$k$ controlling the fraction of highest-magnitude entries
retained in~$\tau$, yielding a two-dimensional genotype~$(\lambda, k)$.
TIES was originally designed to resolve sign conflicts among multiple task
vectors; with only two checkpoints, however, there can be no conflict to
resolve, so its sign-election step becomes trivial and the only effective
modification over TA is the magnitude-based trimming of~$\tau$.
(ii)~An unconstrained \textbf{linear combination}
$\theta_M = \omega_{\mathrm{S2}}\,\theta_{\mathrm{S2}} +
\omega_{\mathrm{S1}}\,\theta_{\mathrm{S1}}$, which removes the
convex-combination constraint.
Despite the richer search spaces of these operators, TA consistently yields the
most robust results across model scales (\Cref{sec:baselines}); we therefore
adopt it as our primary merging strategy.

\section{Method}
\label{sec:method}

\subsection{Multi-objective Evolutionary Search}

\paragraph{Motivation for Pareto Optimization.} As established in our formulation, accuracy and output length are inherently conflicting objectives. Merging configurations that preserve the deeper reasoning behavior of a System 2 model naturally tend to retain verbose chain-of-thought traces, whereas configurations that produce shorter outputs typically incur a loss in symbolic accuracy. Rather than committing to a single, scalarized hyperparameter configuration that forces a premature compromise, Evo-L2S explicitly searches for a Pareto set of solutions. This yields a diverse family of merged models, each representing a distinct accuracy-length trade-off, allowing practitioners to select the operating point that best matches their specific efficiency-performance constraints after downstream evaluation.

\paragraph{Fitness function.} Given a candidate merged model $M$ and a fixed evaluation subset $\mathcal{S}$, we compute two quantities:
\begin{align}
\text{Acc}(M) &= \frac{1}{|\mathcal{S}|}\sum_{i\in\mathcal{S}}\mathbf{1}[c_{i}=1], \\
\text{Len}(M) &= \frac{1}{|\mathcal{S}|}\sum_{i\in\mathcal{S}}l_{i},
\label{eq:fitness}
\end{align}
where $c_{i}\in\{0,1\}$ indicates whether $M$ correctly solves item $i$ and $l_{i}$ is the number of tokens in the corresponding output. Following the minimization convention of PyMoo, the bi-objective fitness vector is $F(M)=[-\text{Acc}(M), \text{Len}(M)]$, so that minimizing $F$ simultaneously maximizes accuracy and minimizes output length.

\paragraph{Evolutionary Algorithm.} We approximate the Pareto front using NSGA-II \citep{996017}, a widely adopted elitist multi-objective evolutionary algorithm. Assuming all objectives are to be minimized, a solution $x_1$ Pareto-dominates $x_2$ (written $x_1 \prec x_2$) if $x_1$ is no greater than $x_2$ on every objective and strictly smaller on at least one; solutions that are not dominated by any other candidate form the Pareto front.

Starting from a population of $N$ genotypes sampled uniformly at random, NSGA-II iterates for $T$ generations: at each step, offspring are produced via binary tournament selection, Simulated Binary Crossover (SBX), and Polynomial Mutation; the parent and offspring populations are then pooled and partitioned into successive non-dominated fronts; the next generation is filled by adding fronts in rank order, breaking ties within the last admissible front by crowding distance: a density measure that favors solutions located in sparser regions of the front, so as to promote a well-spread set of trade-offs. After $T$ generations the final non-dominated set is returned as an approximation of the Pareto front.

\subsection{Subset-Based Fitness Estimation}\label{sec:fitness}

Evolutionary search requires evaluating hundreds of candidate models. Assessing every individual on a full benchmark at each generation is computationally intractable. Evo-L2S therefore approximates full-benchmark performance using a compact subset $\mathcal{S}$ of 50 reasoning problems sampled once from MATH \citep{hendrycksmath2021} and held fixed, which do not overlap with the evaluation test sets. This ensures the fitness signal remains directly comparable across candidates and consistently guides the evolutionary search.

\begin{figure}[t]
  \centering
  \includegraphics[width=\linewidth]{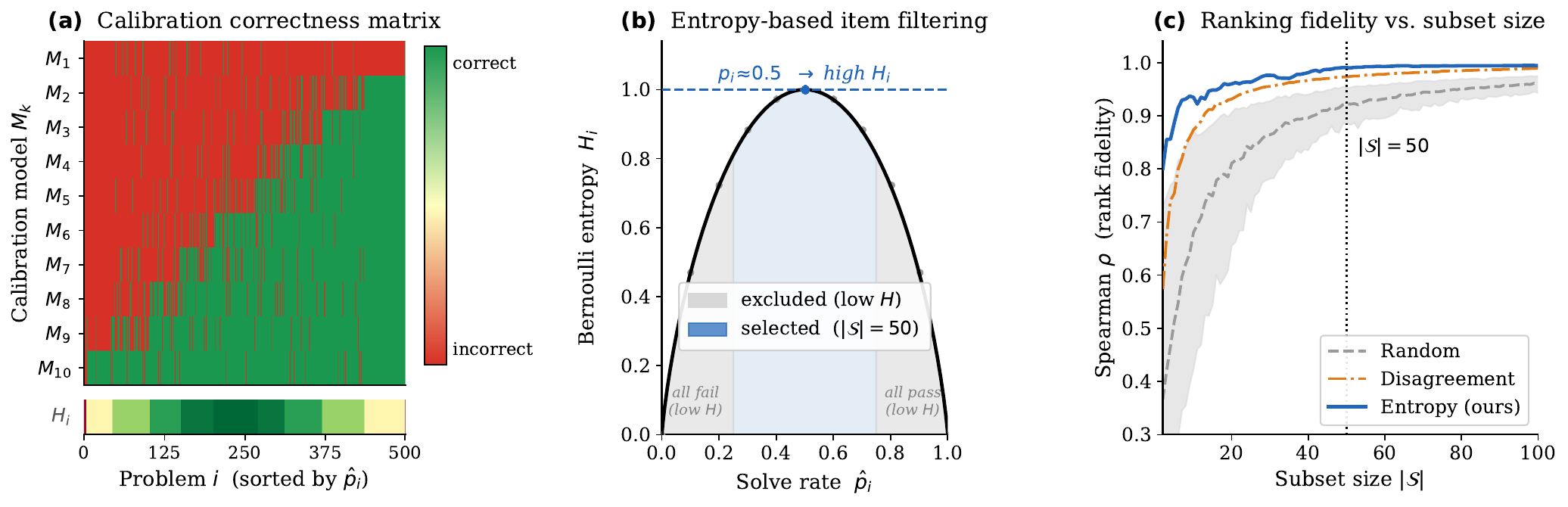}
  \caption{%
    \textbf{Entropy-based subset sampling for simulated ranking.}
    \textbf{(a)}~Calibration correctness matrix: each row is one of $K=10$ candidates, each column a problem sorted by empirical solve rate~$\hat{p}_i$; the colour strip below shows the corresponding Bernoulli entropy~$H_i$. We generate this matrix simulating common patterns observed in correctness matrices, in order to avoid test contamination. \textbf{(b)}~Entropy curve $H_i = -p_i\log_2 p_i - (1{-}p_i)\log_2(1{-}p_i)$; individual problems are scattered at $(\hat{p}_i, H_i)$, with selected items ($H_i$ above the dashed threshold) shown in blue. \textbf{(c)}~Spearman rank correlation~$\rho$ between subset-induced and full-benchmark model rankings as a function of subset size~$|\mathcal{S}|$. Entropy sampling reaches high fidelity at $|\mathcal{S}|=50$, clearly outperforming the baselines.}
  \label{fig:entropy_sampling}
\end{figure}

\paragraph{Entropy sampling.} Not all items are equally useful for ranking candidates. Items solved by every model, or failed by every model, carry no discriminative signal. To construct an optimally informative subset $\mathcal{S}$, we evaluate a calibration pool of $K=10$ merged checkpoints (obtained by uniformly spacing $\lambda$ across $[0,1]$; see \Cref{fig:entropy_sampling}a).

For each item $i$, we compute its empirical correctness probability $p_{i}=\frac{1}{K}\sum_{k=1}^{K}c_{i,k}$, where $c_{i,k}\in\{0,1\}$ indicates whether model $M_{k}$ correctly solved instance $i$. We then quantify the informativeness of each instance using its Bernoulli entropy (\Cref{fig:entropy_sampling}b):
\begin{equation}
H_{i}=-p_{i}\log_{2}p_{i}-(1-p_{i})\log_{2}(1-p_{i})
\end{equation}
Entropy is maximized when $p_{i}\approx0.5$ (i.e., maximal disagreement across candidate models). We rank items by $H_{i}$ and select the top $|\mathcal{S}|=50$ to use as the fixed evaluation subset.

\paragraph{Why high-entropy items?} Intuitively, a problem solved by every model (or by none) carries no signal for ranking candidates; only problems where models disagree reveal meaningful differences. High-entropy items, those with $p_{i}\approx0.5$, are precisely the ones that maximally separate candidates in expectation. Formally, under a simple threshold model of item difficulty, maximizing the expected number of pairwise distinctions between candidate models is mathematically equivalent to selecting items based on their Bernoulli entropy. We refer the interested reader to \Cref{app:entropy} for the complete theoretical derivation.

\paragraph{Ranking fidelity and Baselines.} To validate this theoretical intuition, we compare our entropy sampling approach against two distinct baselines: uniform random sampling and disagreement sampling, the latter selectively retaining only those evaluation items where the extreme endpoints $\lambda\in\{0,1\}$ yield strictly diverging correctness outcomes.

Because NSGA-II relies on ranking-based selection, a proxy subset is effective if the relative ordering of candidates it induces matches the ordering obtained from the full benchmark. We quantify this agreement using the Spearman rank correlation $\rho$. A value of $\rho\approx1$ indicates that the subset faithfully reproduces the full-benchmark ordering. Our proposed entropy sampling approach achieves the highest $\rho$ in the low-budget regime, reaching near-perfect rank agreement much faster than the alternative baseline strategies (\Cref{fig:entropy_sampling}c).

\section{Experiments and Analysis}
\label{sec:experiments}

\subsection{Evaluation Protocol}

\paragraph{Mathematical Reasoning Benchmarks.}
We evaluate all models on six widely used mathematical benchmarks spanning a broad range of difficulty and abstraction: \texttt{GSM8K}~\citep{DBLP:journals/corr/abs-2110-14168} ($1319$ problems),
\texttt{MATH500}~\citep{lightman2024lets} (500), \texttt{Minerva-Math}~\citep{NEURIPS2022_18abbeef} ($272$), \texttt{OlympiadBench}~\citep{he-etal-2024-olympiadbench} ($675$), \texttt{College-Math}~\citep{tang2024mathscale} ($2818$),
and \texttt{AIME24}~\citep{aime24} ($30$).

\paragraph{Prompting and decoding strategies.}
All models are evaluated with \texttt{vLLM}~\citep{kwon2023efficient} in \texttt{bfloat16} precision with a fixed random seed of $0$, using the QwenLM evaluation toolkit, which provides standardized prompt templates and automated procedures for answer extraction and parsing across all benchmarks.

Fast-thinking models are evaluated in a few-shot chain-of-thought setting, where reasoning behavior is elicited via Question--Answer demonstrations included in the prompt. Slow-thinking models are evaluated zero-shot, with an explicit instruction to ``reason step by step'' and enclose the final answer in a \texttt{\textbackslash boxed\{\}} expression. Both settings use greedy decoding (\texttt{temperature}~$= 0.0$, \texttt{top-}$p = 1.0$), with a maximum of $8,192$ new tokens for fast-thinking models and $10,240$ for slow-thinking ones.

All merged models are evaluated under the same configuration as the slow-thinking baseline, so that any change in accuracy or response length can be attributed solely to the merging procedure. Response length is measured as the number of tokens generated per output, using the model's own tokenizer, before any post-processing.

\subsection{Validating entropy-based sampling}
\label{sec:entropy_sampling_validation}

We now validate entropy-based sampling using actual model evaluations rather than a simulated fitness landscape. Specifically, we evaluate ten 7B merged checkpoints, with interpolation coefficients uniformly spaced in $[0,1]$, on the complete set of benchmarks. For each of 30 independent resampling runs, we draw without replacement a random calibration pool containing 90\% of the MATH items. Within this pool, each sampling method selects 50 items. We rank the ten calibration checkpoints using their accuracy on this subset and compute Spearman's $\rho$ against their ranking on each complete target benchmark. \Cref{tab:spearman_7B_math_to_all_5pct} reports the mean correlation and the corresponding 95\% confidence-interval half-width.

Entropy-based sampling yields the highest average rank correlation, reaching $77.2$, compared with $76.8$ for disagreement sampling and $72.2$ for random sampling. This indicates that a small entropy-selected subset of MATH can preserve the relative ranking of merged checkpoints across related mathematical reasoning benchmarks. The lower correlations on College Math show that this transfer is not uniform across all datasets, as confirmed by the results for the 1.5B models (\Cref{tab:spearman_1.5B_math_to_all_5pct}). Overall, the evidence supports the use of entropy-based subset fitness as a computationally efficient proxy for guiding the evolutionary search.

\subsection{Baselines}
\label{sec:baselines}

\paragraph{Reference endpoints.}
\texttt{DeepSeek-R1-Distill-Qwen} (System~2) substantially outperforms \texttt{Qwen2.5-Math} (System~1) at both 1.5B (45.5\% vs.\ 27.6\%) and 7B (59.2\% vs.\ 32.6\%), but generates responses roughly $5\times$ longer. The two models therefore define opposite ends of the accuracy--length trade-off (\Cref{fig:pareto}), motivating the search for intermediate operating points while preserving model capabilities.

\paragraph{Fixed-configuration merging.}
Following \cite{wu2025unlockingefficientlongtoshortllm}, we evaluate Average Merging, Task Arithmetic (TA), and TIES-Merging. Their performance varies considerably across scales: at 1.5B (\Cref{tab:1.5B_res}), Average Merging matches System~2 accuracy while reducing length by 40\%, whereas TA loses 4.6~pp; at 7B (\Cref{tab:7b_res}), TA performs best (-0.7~pp, 38\% reduction), while Average Merging loses 4.5~pp. We also evaluate ACM~\citep{yao2025activationguided}, which computes layer-wise coefficients from an arithmetic pre-merge. Its performance remains sensitive to this initialization, with accuracy spreads of 4.1~pp at 1.5B and 5.6~pp at 7B across pre-merges. Overall, no fixed configuration consistently achieves both high accuracy and substantial compression across scales.

\paragraph{Search baselines.}
We compare Evo-L2S against both static and single-objective search strategies. For the 7B Task Arithmetic space, grid and uniform random search use the same candidate-evaluation budget as Evo-L2S, with the same non-dominated sorting applied to all evaluated candidate sets (\Cref{tab:7b_res}). Random search is considerably stronger than grid search, reaching 54.6\% average accuracy with a 65.8\% average length reduction, but does not fully recover the higher-accuracy region found by Evo-L2S. For \Cref{fig:pareto}, we run evolutionary search with a single objective: optimizing accuracy preserves System~2 performance but yields limited compression ($\sim$38\%), while optimizing length reduces outputs by 71--78\% at the cost of up to 11.6~pp in accuracy. These extremes motivate jointly optimizing both objectives. At 7B scale, evolutionary search with a random subset (Evo-L2S Random in \Cref{fig:pareto}) mainly contributes solutions in the higher-compression, lower-accuracy region, whereas entropy sampling better covers the high-accuracy portion of the Pareto front.

\begin{table}[t!]
  \centering
  \resizebox{\textwidth}{!}{%
    \begin{tabular}{lcccccccc}
      \toprule[0.8pt]
      \textbf{Method} & \textbf{AIME 2024} & \textbf{College Math} & \textbf{GSM8K} & \textbf{MATH-500} & \textbf{Minerva Math} & \textbf{OlympiadBench} & \textbf{MATH} & \textbf{Average} \\
      \midrule
      Entropy & $\mathbf{84.7\pm1.0}$ & $13.7\pm3.0$ & $87.1\pm2.0$ & $\mathbf{90.6\pm1.5}$ & $\mathbf{82.4\pm2.5}$ & $\mathbf{89.7\pm1.3}$ & $\mathbf{92.4\pm0.8}$ & $\mathbf{77.2}$ \\
      Disagreement & $80.6\pm2.2$ & $\mathbf{21.3\pm3.7}$ & $\mathbf{87.6\pm1.8}$ & $88.8\pm1.8$ & $79.1\pm2.7$ & $89.1\pm1.7$ & $91.1\pm2.1$ & $76.8$ \\
      Random & $76.1\pm4.3$ & $19.1\pm5.3$ & $81.0\pm3.8$ & $83.9\pm2.7$ & $75.6\pm2.9$ & $83.8\pm3.2$ & $85.6\pm3.4$ & $72.2$ \\
      \bottomrule[0.8pt]
    \end{tabular}
  }
  \caption{\textbf{Validation of entropy-based sampling for actual rankings.}
  We report Spearman's rank correlation between 7B model rankings estimated from 5\% of MATH and rankings on each full benchmark. Correlation values are computed as $100\rho$; uncertainties are 95\% confidence-interval half-widths across 30 calibration-pool resamples.}
  \label{tab:spearman_7B_math_to_all_5pct}
\end{table}

\subsection{Multi-objective evolutionary merging}

We evaluate the full Evo-L2S pipeline by comparing entropy-based subset selection against uniform random sampling. \Cref{fig:pareto} shows the resulting Pareto fronts at the 1.5B and 7B scales, while complete results for all three model sizes are reported in \Cref{app:results}.

At 1.5B and 7B, Evo-L2S recovers useful intermediate operating points that preserve strong accuracy while substantially reducing reasoning length.
At 1.5B, several solutions match or exceed the System~2 baseline with roughly 50--60\% shorter outputs. At 7B, entropy sampling recovers a broad high-accuracy frontier, spanning from near-baseline accuracy at moderate compression to substantially shorter models at a gradual accuracy cost. Evolutionary Search with a random subset (Evo-L2S Random) mainly contributes solutions in the higher-compression, lower-accuracy region, suggesting that entropy-based fitness estimation more effectively guides the search toward useful trade-offs.

\paragraph{Compression across model scales.}
The 14B results reveal a different compression regime. Its System~2 baseline already produces shorter reasoning traces than the 7B counterpart, suggesting less overthinking and therefore less redundant computation available for removal. Consistently, the 14B Pareto front is considerably steeper: further shortening the reasoning trace affects accuracy more rapidly. This suggests that L2S effectiveness depends not simply on model size, but on how much redundant reasoning is present in the original model.

\begin{figure}[t]
  \centering
  \includegraphics[width=\linewidth]{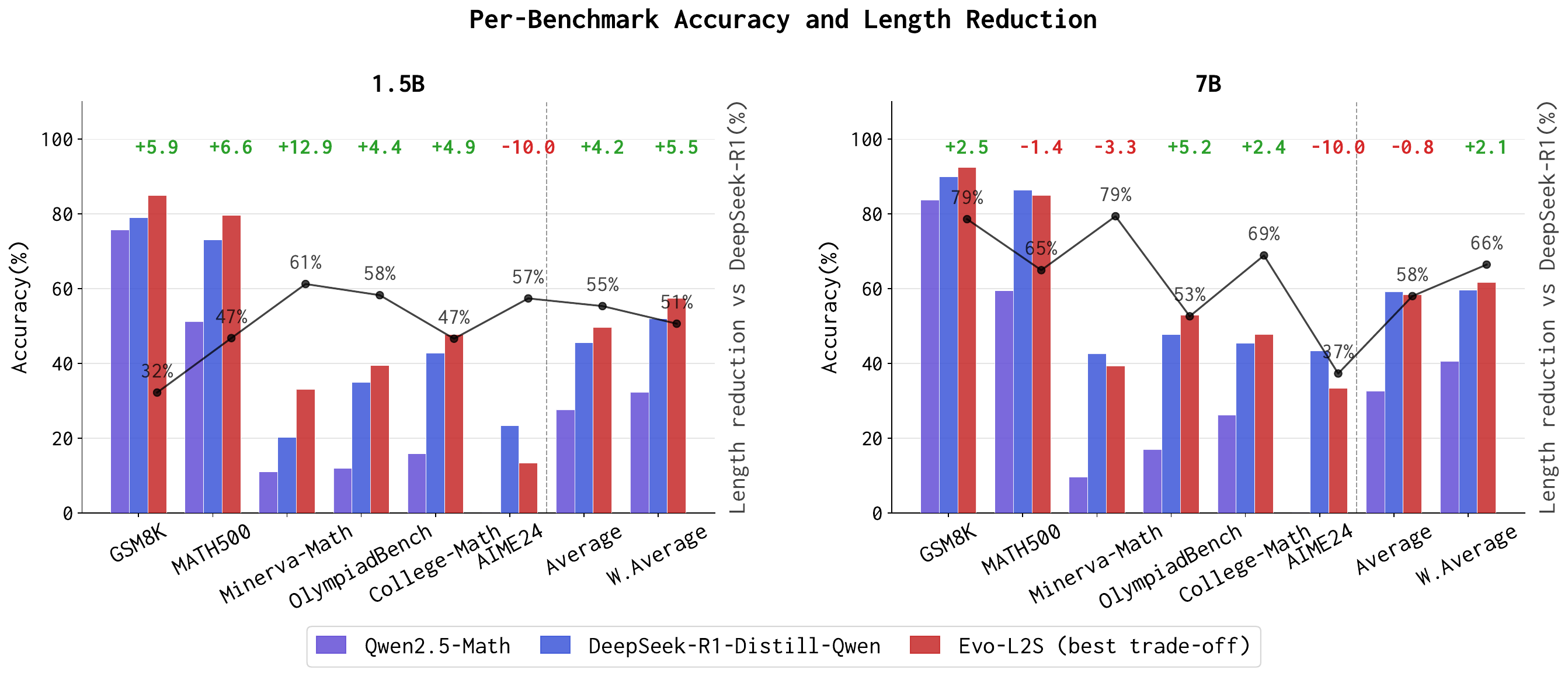}
  \caption{Benchmark-wise accuracy and length reduction of Evo-L2S (best
  trade-off) vs.\ \texttt{DeepSeek-R1-Distill-Qwen} at the \textbf{1.5B} (left) and
  \textbf{7B} (right) scales. Annotations above Evo-L2S bars indicate
  accuracy difference ({\color{green!50!black} improvement} or {\color{red!70!black} degradation}) relative to \texttt{DeepSeek-R1}. The line reports per-benchmark and average output-length reduction (\%).}
  \label{fig:benchmark}
\end{figure}

\subsection{Analysis}

\paragraph{Cross-benchmark behavior.}
\Cref{fig:benchmark} compares a representative high-accuracy,
high-compression Evo-L2S solution against \texttt{DeepSeek-R1-Distill-Qwen} across the six evaluation benchmarks. At 1.5B, Evo-L2S improves accuracy on five of the six benchmarks while reducing response length by 55\% on average. The largest gains occur on Minerva-Math (+12.9 pp), MATH500 (+6.6 pp), and GSM8K (+5.9 pp). At 7B, the stronger System~2 baseline leaves less room for improvement: Evo-L2S gains accuracy on GSM8K, OlympiadBench, and College-Math, while incurring small losses on MATH500 and Minerva-Math. Nevertheless, average accuracy decreases by only 0.8 pp while reasoning length is reduced by 58\%. AIME24 is the main exception at both scales, although its 10 pp difference corresponds to only three of the benchmark's 30 problems.

\paragraph{Compression across model scales.}
The Pareto-front at 14B (\Cref{tab:14B_res}) provides additional insight into
when L2S merging is most effective. The 14B System~2 model already produces
shorter reasoning traces than its 7B counterpart, suggesting less
overthinking and therefore less redundant computation available for removal.
Consistently, its Pareto frontier is steeper: further shortening the
reasoning trace affects accuracy more rapidly. This indicates that the
benefit of L2S compression depends not only on model size, but also on the
amount of redundant reasoning present in the original reasoning model.

\paragraph{Qualitative behavior.}
Finally, the example in \Cref{app:qualitative} illustrates what this
compression can look like at the level of individual reasoning traces.
Rather than simply truncating the solution, Evo-L2S can remove repeated
verification and redundant deliberation while retaining the correct
reasoning trajectory.

\section{Conclusion}

In this work, we introduced \textbf{Evo-L2S}, a novel framework formulating the Long-to-Short (L2S) reasoning task as a multi-objective optimization problem. Evo-L2S demonstrates that evolutionary model merging effectively navigates the complex Pareto frontier between accuracy and output length, overcoming the brittleness of fixed-hyperparameter arithmetic methods that force premature and suboptimal compromises.

To make this search computationally tractable, we introduced a theoretically grounded, entropy-based subset sampling technique that drastically reduces fitness estimation overhead. Evo-L2S reduces reasoning length by over 50\% at the 1.5B and 7B scales while preserving or improving problem-solving accuracy; at 14B, the steeper Pareto front reveals that attainable compression depends on the intrinsic compressibility of the reasoning model.

Ultimately, Evo-L2S provides a deployable solution to the inference bottlenecks of reasoning models. By decoupling System 2 reasoning capabilities from their verbosity, practitioners can dynamically select Pareto-optimal models meeting specific latency and compute constraints. This demonstrates that weight-space evolutionary optimization is a highly effective, training-free mechanism to align generation length with strict efficiency requirements.

\section*{Acknowledgments}
This work was supported by Sapienza University of Rome via the Startup Research grant ``TVT'', and by MUR FIS2 grant n.\ FIS-2023-00942 ``NEXUS'' (cup B53C25001030001). We also acknowledge ISCRA for awarding this project access to the LEONARDO supercomputer, owned by the EuroHPC Joint Undertaking, hosted by CINECA (Italy).

\bibliography{colm2026_conference}
\bibliographystyle{colm2026_conference}

\appendix

\clearpage
\section{Limitations}
Our generation procedure is deterministic, but the evolutionary search is reported from one optimization seed. Consequently, the present experiments do not quantify variation of the final Pareto front across NSGA-II initializations. The calibration-pool resampling analysis in Section~\ref{sec:entropy_sampling_validation} evaluates the stability of the fitness proxy, not the stability of the evolutionary algorithm itself. Further analysis of search stability is left for future work.

\section{Why entropy selection is informative}
\label{app:entropy}
The following proposition gives a simple justification for entropy-based subset
selection.

\textbf{Proposition.}
Assume candidate merged models are indexed by $\lambda \in [0,1]$, and that for
each problem $i$ there exists a threshold $t_i \in [0,1]$ such that the model
solves the problem iff $\lambda \ge t_i$, i.e.
\[
c_i(\lambda) = \mathbf{1}\{\lambda \ge t_i\}.
\]
Let $\lambda,\lambda' \overset{\mathrm{i.i.d.}}{\sim} \mathrm{Unif}[0,1]$, and define
\[
D_i(\lambda,\lambda') = \mathbf{1}\{c_i(\lambda) \neq c_i(\lambda')\},
\]
the indicator that problem $i$ distinguishes the two candidate models. Then
\[
\mathbb{E}_{\lambda,\lambda'}[D_i] = 2 p_i(1-p_i),
\]
where
\[
p_i = \mathbb{P}_{\lambda}(c_i(\lambda)=1).
\]
Consequently, for any subset budget $m$, the subset of $m$ problems that
maximizes the expected number of pairwise distinctions,
\[
\mathbb{E}_{\lambda,\lambda'}\!\left[\sum_{i \in \mathcal S} D_i(\lambda,\lambda')\right],
\]
is obtained by selecting the $m$ problems with largest Bernoulli entropy
\[
H_i = -p_i \log_2 p_i - (1-p_i)\log_2(1-p_i).
\]

\textit{Proof.}
Under the threshold model, $D_i(\lambda,\lambda')=1$ iff one of $\lambda,\lambda'$
lies below $t_i$ and the other lies above it. Therefore
\[
\mathbb{E}[D_i]
= \mathbb{P}(\lambda < t_i,\lambda' \ge t_i)
 + \mathbb{P}(\lambda' < t_i,\lambda \ge t_i)
= t_i(1-t_i) + (1-t_i)t_i
= 2t_i(1-t_i).
\]
Since
\[
p_i = \mathbb{P}_{\lambda}(c_i(\lambda)=1) = 1-t_i,
\]
we obtain
\[
\mathbb{E}[D_i] = 2p_i(1-p_i).
\]
Hence the expected total number of pairwise distinctions over a subset
$\mathcal S$ is
\[
\sum_{i \in \mathcal S} 2p_i(1-p_i),
\]
so the optimal subset consists of the items with the largest values of
$p_i(1-p_i)$. Finally, both $p \mapsto p(1-p)$ and the Bernoulli entropy
$p \mapsto -p\log p -(1-p)\log(1-p)$ are symmetric around $1/2$ and strictly
increasing on $[0,1/2]$, so they induce the same ranking of items. $\square$

\textit{Remark.}
The threshold assumption is stylized, but the same conclusion holds locally for
smooth item-response curves. If $q_i(\lambda)=\mathbb{P}(c_i(\lambda)=1)$ is
differentiable and follows a logistic form
$q_i(\lambda)=\sigma(a_i(\lambda-b_i))$, then for nearby candidates
$\lambda$ and $\lambda+\delta$,
\[
q_i(\lambda+\delta)-q_i(\lambda)
= a_i\,q_i(\lambda)\bigl(1-q_i(\lambda)\bigr)\,\delta + O(\delta^2).
\]
Thus, for comparable slopes $a_i$, items with $q_i(\lambda)\approx 1/2$ are
again the most sensitive to changes in the merge coefficient, which is exactly
what entropy sampling favors.

\section{Reproducibility}
To facilitate future research and ensure the full reproducibility of our multi-objective evolutionary merging framework, we publicly release our complete codebase at the following link: \url{https://github.com/iacobellimario/L2S-MultiObjective-Evolutionary-Merging}.

\paragraph{Models and Frameworks} 
All base checkpoints in our experiments are publicly available via Hugging Face. We utilized the \texttt{Qwen2.5-Math} (System 1) and \texttt{DeepSeek-R1-Distill-Qwen} (System 2) model families at the 1.5B, 7B, and 14B parameter scales. Parameter-space merging operations were executed using the \texttt{MergeKit} library (Goddard et al., 2024), while the evolutionary optimization and Pareto front approximations were driven by \texttt{PyMoo} (Blank \& Deb, 2020). Inference and generation were conducted using \texttt{vLLM} (Kwon et al., 2023) in \texttt{bfloat16} precision to maximize throughput during fitness evaluation.

\paragraph{Evolutionary Search Hyperparameters} 
The NSGA-II algorithm was initialized with a population size of $N=20$ candidates and executed for $T=10$ generations. The evolutionary operators utilized were Simulated Binary Crossover (SBX) and Polynomial Mutation. Fitness estimation during the search was performed on a static subset $\mathcal{S}$ of 50 problems sampled from the MATH dataset. This subset was deterministically selected via our entropy-based sampling method from a calibration pool of $K=10$ checkpoints, which were uniformly spaced along the Task Arithmetic interpolation line ($\lambda \in [0,1]$). 

\paragraph{Evaluation Protocol} 
All formal baseline and Pareto-front evaluations were performed using the official \texttt{QwenLM} evaluation toolkit to guarantee standardized prompt templates and unbiased answer extraction. Generation hyperparameters were strictly controlled across all tested models, utilizing greedy decoding ($\text{temperature} = 0.0, \text{top-}p = 1.0$) with maximum generation limits of 8,192 tokens for System 1 models and 10,240 tokens for System 2 models. A fixed random seed of 0 was enforced across the entire pipeline to ensure deterministic execution.

\paragraph{Compute Infrastructure}
The evolutionary search and all subsequent evaluations were executed on a cluster equipped with NVIDIA A100 (64GB) GPUs. The evolutionary search for the 1.5B and 7B searches require 7.5 and 14.5 A100 GPU-hours, respectively, while the 14B search runs for 24 hours on two A100s (48 GPU-hours). This is a one-time offline optimization cost; the deployed merged checkpoint uses standard decoding and introduces no additional per-token controller or search procedure.

\begin{table}[t]
\centering
\begin{tabular}{lccc}
\toprule
 & \textbf{1.5B} & \textbf{7B} & \textbf{14B} \\
\midrule
Population size              & 20       & 20        & 12 \\
Generations                  & 10       & 10        & 8  \\
Candidate evaluations        & 200      & 200       & 96 \\
Items per evaluation         & 50       & 50        & 50 \\
Approx.\ forward passes      & 10{,}000 & 10{,}000  & 4{,}800 \\
A100 GPU-hours               & 7.5      & 14.5      & 48 \\
\bottomrule
\end{tabular}
\caption{Evolutionary-search compute cost by model scale.}
\label{tab:search_cost}
\end{table}

\begin{table}[t]
  \centering
  \resizebox{\textwidth}{!}{%
    \begin{tabular}{lcccccccc}
      \toprule[0.8pt]
      \textbf{Method} & \textbf{AIME 2024} & \textbf{College Math} & \textbf{GSM8K} & \textbf{MATH-500} & \textbf{Minerva Math} & \textbf{OlympiadBench} & \textbf{MATH} & \textbf{Average} \\
      \midrule
      Entropy & $\mathbf{49.7\pm3.8}$ & $-9.5\pm1.5$ & $\mathbf{82.8\pm1.1}$ & $\mathbf{90.7\pm1.2}$ & $29.3\pm2.4$ & $\mathbf{90.1\pm1.1}$ & $\mathbf{90.4\pm1.3}$ & $\mathbf{60.5}$ \\
      Disagreement & $38.4\pm6.7$ & $0.4\pm7.0$ & $81.1\pm3.3$ & $83.9\pm3.6$ & $34.4\pm6.1$ & $83.9\pm3.1$ & $83.9\pm3.7$ & $58.0$ \\
      Random & $34.2\pm6.6$ & $\mathbf{1.2\pm7.3}$ & $76.2\pm5.4$ & $77.7\pm5.7$ & $\mathbf{34.5\pm8.2}$ & $78.4\pm5.9$ & $77.6\pm5.9$ & $54.3$ \\
      \bottomrule[0.8pt]
    \end{tabular}%
  }
  \caption{Spearman rank correlation between 1.5B model rankings estimated from 5\% of MATH and rankings on each full target benchmark. Values are reported as $100\rho$; uncertainties are 95\% confidence-interval half-widths across 30 calibration-pool resamples.}
  \label{tab:spearman_1.5B_math_to_all_5pct}
\end{table}

\section{Detailed Results}
\label{app:results}

In \Cref{tab:1.5B_res,tab:7b_res,tab:14B_res}, we report complete numerical results for all three model scales (1.5B, 7B, and 14B) across six mathematical reasoning benchmarks (GSM8K, MATH500, Minerva-Math, OlympiadBench, College-Math, AIME24), together with arithmetic average and weighted average (weighted by benchmark size). Each entry reports accuracy (\%) in the top row and output length (tokens) in the bottom row; for the Average and Weighted Average  columns, the bottom row instead reports length reduction relative to \texttt{DeepSeek-R1-Distill-Qwen} [\%]. The highest accuracy in each column is highlighted in \textbf{bold}. Evo-L2S rows correspond to entropy-sampling Pareto-optimal solutions, sorted by decreasing average accuracy.

\begin{table}[ht]
    \centering
    \def\arraystretch{1.2}
    \resizebox{\textwidth}{!}{%
    \begin{tabular}{lcccccccc}
    \toprule[0.8pt]
       \diagbox{Method}{Bench} & GSM8K & MATH500
       & \makecell[c]{Minerva\\Math} & \makecell[c]{Olympiad\\Bench} & \makecell[c]{College\\Math} & AIME24 & Avg. & W-Avg. \\
       \hline
       \multirow{2}{*}{\makecell[l]{Qwen2.5-Math-1.5B}}
       & 75.7 & 51.2 & 11.0 & 11.9 & 15.8 & 0.0 & 27.6 & 32.3 \\
       & \cellcolor{lightblue}(110.0) & \cellcolor{lightblue}(327.2) & \cellcolor{lightblue}(1032.2) & \cellcolor{lightblue}(1337.7) & \cellcolor{lightblue}(606.3) & \cellcolor{lightblue}(1107.5) & \cellcolor{lightblue}(753.5) & \cellcolor{lightblue}(576.1) \\
       \multirow{2}{*}{\makecell[l]{DeepSeek-R1-Distill-Qwen-1.5B}}
       & 79.0 & 73.0 & 20.2 & 35.0 & 42.8 & \textbf{23.3} & 45.5 & 51.9 \\
       & \cellcolor{lightblue}(628.7) & \cellcolor{lightblue}(2570.0) & \cellcolor{lightblue}(3014.8) & \cellcolor{lightblue}(5814.2) & \cellcolor{lightblue}(1776.6) & \cellcolor{lightblue}(8432.9) & \cellcolor{lightblue}(3706.2) & \cellcolor{lightblue}(2158.6) \\
       \hline
       \rowcolor{lightgray}
       \multicolumn{9}{c}{\textit{Arithmetic Merging}} \\
       \multirow{2}{*}{Average Merging}
       & 78.2 & 71.4 & 30.1 & 36.4 & 43.6 & 13.3 & 45.5 & 52.5 \\
       & \cellcolor{lightblue}(533.1) & \cellcolor{lightblue}(1651.0) & \cellcolor{lightblue}(1657.1) & \cellcolor{lightblue}(3057.5) & \cellcolor{lightblue}(1117.6) & \cellcolor{lightblue}(5301.2) & \cellcolor{lightblue}[40.1\%] & \cellcolor{lightblue}[39.3\%] \\
       \multirow{2}{*}{Task Arithmetic (TA)}
       & 75.0 & 66.6 & 27.2 & 31.1 & 38.5 & 6.7 & 40.9 & 48.0 \\
       & \cellcolor{lightblue}(1219.2) & \cellcolor{lightblue}(2731.9) & \cellcolor{lightblue}(2638.1) & \cellcolor{lightblue}(5186.3) & \cellcolor{lightblue}(1634.4) & \cellcolor{lightblue}(6564.3) & \cellcolor{lightblue}[10.2\%] & \cellcolor{lightblue}[1.0\%] \\
       \multirow{2}{*}{TIES-Merging}
       & 78.8 & 72.2 & 31.6 & 33.6 & 44.7 & 6.7 & 44.6 & 53.0 \\
       & \cellcolor{lightblue}(595.8) & \cellcolor{lightblue}(1757.5) & \cellcolor{lightblue}(1733.2) & \cellcolor{lightblue}(3828.8) & \cellcolor{lightblue}(1101.6) & \cellcolor{lightblue}(4692.1) & \cellcolor{lightblue}[38.4\%] & \cellcolor{lightblue}[34.3\%] \\
       \hline
       \rowcolor{lightgray}
       \multicolumn{9}{c}{\textit{Activation-informed Merging (ACM)}} \\
       \multirow{2}{*}{ACM-Average}
       & 75.1 & 69.0 & 30.1 & 33.5 & 42.8 & 3.3 & 42.3 & 50.8 \\
       & \cellcolor{lightblue}(502.8) & \cellcolor{lightblue}(1370.6) & \cellcolor{lightblue}(1026.5) & \cellcolor{lightblue}(2539.7) & \cellcolor{lightblue}(950.0) & \cellcolor{lightblue}(4182.6) & \cellcolor{lightblue}[52.5\%] & \cellcolor{lightblue}[49.2\%] \\
       \multirow{2}{*}{ACM-TA}
       & 80.2 & 73.0 & \textbf{33.5} & 36.3 & 42.3 & 13.3 & 46.4 & 52.6 \\
       & \cellcolor{lightblue}(576.3) & \cellcolor{lightblue}(1835.0) & \cellcolor{lightblue}(1522.3) & \cellcolor{lightblue}(3140.1) & \cellcolor{lightblue}(1183.4) & \cellcolor{lightblue}(5488.4) & \cellcolor{lightblue}[38.2\%] & \cellcolor{lightblue}[36.4\%] \\
       \multirow{2}{*}{ACM-TIES}
       & 78.2 & 69.4 & 27.9 & 32.7 & 44.9 & 13.3 & 44.4 & 52.4 \\
       & \cellcolor{lightblue}(469.8) & \cellcolor{lightblue}(1586.0) & \cellcolor{lightblue}(1263.9) & \cellcolor{lightblue}(2698.9) & \cellcolor{lightblue}(992.1) & \cellcolor{lightblue}(4201.4) & \cellcolor{lightblue}[49.6\%] & \cellcolor{lightblue}[46.4\%] \\
       \hline
       \rowcolor{lightgray}
       \multicolumn{9}{c}{\textit{Single-Objective Evolutionary Merging}} \\
       \multirow{2}{*}{Evo-L2S (Accuracy)}
       & 84.7 & 77.4 & 28.3 & 39.0 & 43.7 & 20.0 & 48.9 & 54.9 \\
       & \cellcolor{lightblue}(543.6) & \cellcolor{lightblue}(1758.2) & \cellcolor{lightblue}(1673.2) & \cellcolor{lightblue}(3651.2) & \cellcolor{lightblue}(1259.4) & \cellcolor{lightblue}(4904.9) & \cellcolor{lightblue}[38.0\%] & \cellcolor{lightblue}[32.2\%] \\
       \multirow{2}{*}{Evo-L2S (Length)}
       & 73.4 & 62.6 & 29.0 & 28.4 & 42.2 & 10.0 & 40.9 & 48.9 \\
       & \cellcolor{lightblue}(388.7) & \cellcolor{lightblue}(782.9) & \cellcolor{lightblue}(853.6) & \cellcolor{lightblue}(1258.4) & \cellcolor{lightblue}(703.3) & \cellcolor{lightblue}(2407.5) & \cellcolor{lightblue}[71.2\%] & \cellcolor{lightblue}[66.7\%] \\
       \hline
       \rowcolor{lightgray}
       \multicolumn{9}{c}{\textit{Multi-Objective Evolutionary Merging (Entropy Sampling, Pareto-optimal)}} \\
       \multirow{2}{*}{Evo-L2S-01}
       & \textbf{85.7} & 77.2 & \textbf{33.5} & \textbf{40.9} & 46.8 & 16.7 & \textbf{50.1} & 57.1 \\
       & \cellcolor{lightblue}(409.5) & \cellcolor{lightblue}(1329.9) & \cellcolor{lightblue}(1059.5) & \cellcolor{lightblue}(2349.8) & \cellcolor{lightblue}(938.8) & \cellcolor{lightblue}(4478.3) & \cellcolor{lightblue}[52.5\%] & \cellcolor{lightblue}[51.6\%] \\
       \multirow{2}{*}{\textbf{Evo-L2S-02}}
       & 84.9 & \textbf{79.6} & 33.1 & 39.4 & \textbf{47.7} & 13.3 & 49.7 & \textbf{57.4} \\
       & \cellcolor{lightblue}(426.2) & \cellcolor{lightblue}(1367.8) & \cellcolor{lightblue}(1168.4) & \cellcolor{lightblue}(2427.1) & \cellcolor{lightblue}(948.5) & \cellcolor{lightblue}(3593.2) & \cellcolor{lightblue}[55.3\%] & \cellcolor{lightblue}[50.6\%] \\
       \multirow{2}{*}{Evo-L2S-03}
       & 81.5 & 70.8 & 30.1 & 34.8 & 46.8 & 3.3 & 44.5 & 54.6 \\
       & \cellcolor{lightblue}(408.7) & \cellcolor{lightblue}(1225.1) & \cellcolor{lightblue}(1044.2) & \cellcolor{lightblue}(2038.5) & \cellcolor{lightblue}(822.2) & \cellcolor{lightblue}(3341.5) & \cellcolor{lightblue}[60.1\%] & \cellcolor{lightblue}[56.9\%] \\
       \multirow{2}{*}{Evo-L2S-04}
       & 73.3 & 61.2 & 27.6 & 29.8 & 42.6 & 10.0 & 40.8 & 49.0 \\
       & \cellcolor{lightblue}(385.5) & \cellcolor{lightblue}(861.3) & \cellcolor{lightblue}(857.9) & \cellcolor{lightblue}(1235.1) & \cellcolor{lightblue}(711.5) & \cellcolor{lightblue}(2770.4) & \cellcolor{lightblue}[69.3\%] & \cellcolor{lightblue}[66.2\%] \\
    \bottomrule[0.8pt]
    \end{tabular}
    }
    \caption{Results of model merging methods at the 1.5B scale.}
    \label{tab:1.5B_res}
\end{table}

\begin{table}[ht]
    \centering
    \def\arraystretch{1.2}
    \resizebox{\textwidth}{!}{%
    \begin{tabular}{lcccccccc}
    \toprule[0.8pt]
       \diagbox{Method}{Bench} & GSM8K & MATH500
       & \makecell[c]{Minerva\\Math} & \makecell[c]{Olympiad\\Bench} & \makecell[c]{College\\Math} & AIME24 & Avg. & W-Avg. \\
       \hline
       \multirow{2}{*}{\makecell[l]{Qwen2.5-Math-7B}}
       & 83.7 & 59.4 & 9.6 & 17.0 & 26.2 & 0.0 & 32.6 & 40.6 \\
       & \cellcolor{lightblue}(104.0) & \cellcolor{lightblue}(432.8) & \cellcolor{lightblue}(946.8) & \cellcolor{lightblue}(1228.2) & \cellcolor{lightblue}(759.4) & \cellcolor{lightblue}(1450.0) & \cellcolor{lightblue}(820.9) & \cellcolor{lightblue}(645.5) \\
       \multirow{2}{*}{\makecell[l]{DeepSeek-R1-Distill-Qwen-7B}}
       & 89.9 & 86.4 & \textbf{42.6} & 47.7 & 45.4 & \textbf{43.3} & 59.2 & 59.6 \\
       & \cellcolor{lightblue}(1794.3) & \cellcolor{lightblue}(3505.1) & \cellcolor{lightblue}(4737.2) & \cellcolor{lightblue}(6187.3) & \cellcolor{lightblue}(3171.1) & \cellcolor{lightblue}(7891.9) & \cellcolor{lightblue}(4547.8) & \cellcolor{lightblue}(3341.1) \\
       \hline
       \rowcolor{lightgray}
       \multicolumn{9}{c}{\textit{Arithmetic Merging}} \\
       \multirow{2}{*}{Average Merging}
       & 91.1 & 83.8 & 39.0 & 47.9 & \textbf{49.5} & 16.7 & 54.7 & 61.5 \\
       & \cellcolor{lightblue}(370.1) & \cellcolor{lightblue}(1098.7) & \cellcolor{lightblue}(951.5) & \cellcolor{lightblue}(2223.1) & \cellcolor{lightblue}(911.1) & \cellcolor{lightblue}(3710.7) & \cellcolor{lightblue}[66.0\%] & \cellcolor{lightblue}[70.8\%] \\
       \multirow{2}{*}{Task Arithmetic}
       & 91.6 & 87.2 & 39.0 & 49.3 & 47.2 & 36.7 & 58.5 & 61.0 \\
       & \cellcolor{lightblue}(634.6) & \cellcolor{lightblue}(1970.4) & \cellcolor{lightblue}(1796.5) & \cellcolor{lightblue}(4304.3) & \cellcolor{lightblue}(1539.1) & \cellcolor{lightblue}(6580.9) & \cellcolor{lightblue}[38.3\%] & \cellcolor{lightblue}[48.0\%] \\
       \multirow{2}{*}{TIES-Merging}
       & 89.9 & 84.0 & 39.0 & 48.7 & 48.5 & 23.3 & 55.6 & 60.8 \\
       & \cellcolor{lightblue}(360.3) & \cellcolor{lightblue}(1296.6) & \cellcolor{lightblue}(1050.6) & \cellcolor{lightblue}(2610.8) & \cellcolor{lightblue}(991.7) & \cellcolor{lightblue}(4266.5) & \cellcolor{lightblue}[61.2\%] & \cellcolor{lightblue}[67.5\%] \\
       \hline
       \rowcolor{lightgray}
       \multicolumn{9}{c}{\textit{Activation-informed Merging (ACM)}} \\
       \multirow{2}{*}{ACM-Average}
       & 87.6 & 81.2 & 37.5 & 41.8 & 47.6 & 20.0 & 52.6 & 58.7 \\
       & \cellcolor{lightblue}(369.6) & \cellcolor{lightblue}(1066.3) & \cellcolor{lightblue}(841.7) & \cellcolor{lightblue}(1981.5) & \cellcolor{lightblue}(862.1) & \cellcolor{lightblue}(2951.5) & \cellcolor{lightblue}[70.4\%] & \cellcolor{lightblue}[72.8\%] \\
       \multirow{2}{*}{ACM-TA}
       & 92.3 & 86.0 & 39.7 & 50.2 & \textbf{49.5} & 30.0 & 57.9 & \textbf{62.3} \\
       & \cellcolor{lightblue}(354.1) & \cellcolor{lightblue}(1157.2) & \cellcolor{lightblue}(942.8) & \cellcolor{lightblue}(2292.2) & \cellcolor{lightblue}(926.5) & \cellcolor{lightblue}(3898.1) & \cellcolor{lightblue}[64.9\%] & \cellcolor{lightblue}[70.3\%] \\
       \multirow{2}{*}{ACM-TIES}
       & 86.1 & 81.6 & 35.7 & 45.0 & 48.9 & 16.7 & 52.3 & 59.3 \\
       & \cellcolor{lightblue}(378.8) & \cellcolor{lightblue}(1054.7) & \cellcolor{lightblue}(1020.8) & \cellcolor{lightblue}(2140.1) & \cellcolor{lightblue}(849.7) & \cellcolor{lightblue}(3118.6) & \cellcolor{lightblue}[68.6\%] & \cellcolor{lightblue}[72.1\%] \\
       \hline
       \rowcolor{lightgray}
       \multicolumn{9}{c}{\textit{Single-Objective Evolutionary Merging}} \\
       \multirow{2}{*}{Evo-L2S (Accuracy)}
       & 92.2 & \textbf{88.0} & 40.4 & 50.7 & 47.3 & 36.7 & 59.2 & 61.5 \\
       & \cellcolor{lightblue}(526.7) & \cellcolor{lightblue}(1792.2) & \cellcolor{lightblue}(1981.5) & \cellcolor{lightblue}(4147.0) & \cellcolor{lightblue}(1492.8) & \cellcolor{lightblue}(6674.3) & \cellcolor{lightblue}[39.1\%] & \cellcolor{lightblue}[50.2\%] \\
       \multirow{2}{*}{Evo-L2S (Length)}
       & 85.1 & 72.8 & 32.4 & 37.0 & 41.9 & 16.7 & 47.6 & 53.6 \\
       & \cellcolor{lightblue}(366.0) & \cellcolor{lightblue}(726.6) & \cellcolor{lightblue}(703.7) & \cellcolor{lightblue}(1295.7) & \cellcolor{lightblue}(762.5) & \cellcolor{lightblue}(2038.9) & \cellcolor{lightblue}[78.4\%] & \cellcolor{lightblue}[78.0\%] \\
       \hline
       \rowcolor{lightgray}
       \multicolumn{9}{c}{
         \textit{Budget-Matched Static Search (Pareto-selected)}
       } \\

       \multirow{2}{*}{Grid Search}
       & 84.5 & 76.0 & 34.2 & 37.4 & 46.7 & 10.0 & 48.1 & 56.3 \\
       & \cellcolor{lightblue}(382.2)
       & \cellcolor{lightblue}(843.2)
       & \cellcolor{lightblue}(725.7)
       & \cellcolor{lightblue}(1348.3)
       & \cellcolor{lightblue}(705.2)
       & \cellcolor{lightblue}(2408.9)
       & \cellcolor{lightblue}[76.5\%]
       & \cellcolor{lightblue}[78.2\%] \\

       \multirow{2}{*}{Random Search}
       & 90.0 & 83.8 & 38.2 & 46.8 & 48.9 & 20.0 & 54.6 & 60.7 \\
       & \cellcolor{lightblue}(347.8)
       & \cellcolor{lightblue}(1187.1)
       & \cellcolor{lightblue}(874.8)
       & \cellcolor{lightblue}(2160.6)
       & \cellcolor{lightblue}(879.4)
       & \cellcolor{lightblue}(3870.5)
       & \cellcolor{lightblue}[65.8\%]
       & \cellcolor{lightblue}[71.5\%] \\
       \hline
       \rowcolor{lightgray}
       \multicolumn{9}{c}{\textit{Multi-Objective Evolutionary Merging (Entropy Sampling, Pareto-optimal)}} \\
       \multirow{2}{*}{Evo-L2S-01}
       & 92.0 & 87.8 & 42.3 & 51.1 & 47.5 & 36.7 & \textbf{59.6} & 61.7 \\
       & \cellcolor{lightblue}(531.2) & \cellcolor{lightblue}(1854.5) & \cellcolor{lightblue}(1824.8) & \cellcolor{lightblue}(4206.5) & \cellcolor{lightblue}(1481.7) & \cellcolor{lightblue}(6674.3) & \cellcolor{lightblue}[39.3\%] & \cellcolor{lightblue}[50.2\%] \\
       \multirow{2}{*}{\textbf{Evo-L2S-02}}
       & \textbf{92.4} & 85.0 & 39.3 & \textbf{52.9} & 47.8 & 33.3 & 58.4 & 61.7 \\
       & \cellcolor{lightblue}(384.0) & \cellcolor{lightblue}(1228.6) & \cellcolor{lightblue}(976.9) & \cellcolor{lightblue}(2932.2) & \cellcolor{lightblue}(986.8) & \cellcolor{lightblue}(4948.4) & \cellcolor{lightblue}[58.0\%] & \cellcolor{lightblue}[66.4\%] \\
       \multirow{2}{*}{Evo-L2S-03}
       & 91.4 & 83.4 & 40.4 & 47.7 & 49.2 & 30.0 & 57.0 & 61.5 \\
       & \cellcolor{lightblue}(362.3) & \cellcolor{lightblue}(1147.3) & \cellcolor{lightblue}(864.3) & \cellcolor{lightblue}(2433.8) & \cellcolor{lightblue}(860.7) & \cellcolor{lightblue}(3140.9) & \cellcolor{lightblue}[67.7\%] & \cellcolor{lightblue}[70.9\%] \\
       \multirow{2}{*}{Evo-L2S-04}
       & 85.4 & 76.4 & 33.8 & 40.7 & 47.1 & 23.3 & 51.1 & 57.2 \\
       & \cellcolor{lightblue}(385.5) & \cellcolor{lightblue}(867.1) & \cellcolor{lightblue}(747.9) & \cellcolor{lightblue}(1448.5) & \cellcolor{lightblue}(700.6) & \cellcolor{lightblue}(3368.4) & \cellcolor{lightblue}[72.4\%] & \cellcolor{lightblue}[77.6\%] \\
       \multirow{2}{*}{Evo-L2S-05}
       & 82.0 & 73.2 & 28.7 & 36.9 & 44.0 & 20.0 & 47.5 & 53.8 \\
       & \cellcolor{lightblue}(378.5) & \cellcolor{lightblue}(867.4) & \cellcolor{lightblue}(731.9) & \cellcolor{lightblue}(1283.2) & \cellcolor{lightblue}(725.7) & \cellcolor{lightblue}(2006.6) & \cellcolor{lightblue}[78.0\%] & \cellcolor{lightblue}[78.1\%] \\
    \bottomrule[0.8pt]
    \end{tabular}
    }
    \caption{Results of model merging methods at the 7B scale. Grid and random search use the same evaluation budget as Evo-L2S. For these static-search baselines, we apply the same non-dominated sorting procedure as Evo-L2S to select a Pareto-optimal checkpoint from the evaluated candidates.}
    \label{tab:7b_res}
\end{table}

\begin{table}[ht]
    \centering
    \def\arraystretch{1.2}
    \resizebox{\textwidth}{!}{%
    \begin{tabular}{lcccccccc}
    \toprule[0.8pt]
       \diagbox{Method}{Bench} & GSM8K & MATH500
       & \makecell[c]{Minerva\\Math} & \makecell[c]{Olympiad\\Bench} & \makecell[c]{College\\Math} & AIME24 & Avg. & W-Avg. \\
       \hline
       \multirow{2}{*}{\makecell[l]{Qwen2.5-14B}}
       & 91.7 & 49.6 & 16.5 & 21.3 & 12.6 & 0.0 & 32.0 & 35.6 \\
       & \cellcolor{lightblue}(202.4) & \cellcolor{lightblue}(598.4) & \cellcolor{lightblue}(1315.8) & \cellcolor{lightblue}(1310.9) & \cellcolor{lightblue}(1231.7) & \cellcolor{lightblue}(1770.0) & \cellcolor{lightblue}(1071.5) & \cellcolor{lightblue}(950.0) \\
       \multirow{2}{*}{\makecell[l]{DeepSeek-R1-Distill-Qwen-14B}}
       & 93.4 & 86.2 & \textbf{45.6} & \textbf{53.5} & 46.1 & \textbf{46.7} & \textbf{61.9} & \textbf{61.7} \\
       & \cellcolor{lightblue}(1061.1) & \cellcolor{lightblue}(3183.6) & \cellcolor{lightblue}(3806.2) & \cellcolor{lightblue}(5765.1) & \cellcolor{lightblue}(2671.1) & \cellcolor{lightblue}(7814.4) & \cellcolor{lightblue}(4050.3) & \cellcolor{lightblue}(2793.0) \\
       \hline
       \rowcolor{lightgray}
       \multicolumn{9}{c}{\textit{Arithmetic Merging}} \\
       \multirow{2}{*}{Average Merging}
       & 89.8 & 81.6 & 39.3 & 44.7 & 45.0 & 33.3 & 55.6 & 58.4 \\
       & \cellcolor{lightblue}(781.8) & \cellcolor{lightblue}(1692.4) & \cellcolor{lightblue}(965.9) & \cellcolor{lightblue}(3227.1) & \cellcolor{lightblue}(1210.7) & \cellcolor{lightblue}(5837.3) & \cellcolor{lightblue}[43.6\%] & \cellcolor{lightblue}[49.6\%] \\
       \multirow{2}{*}{Task Arithmetic}
       & 94.5 & 86.8 & 44.9 & 49.6 & 45.8 & 36.7 & 59.7 & 61.3 \\
       & \cellcolor{lightblue}(865.8) & \cellcolor{lightblue}(2459.2) & \cellcolor{lightblue}(2593.1) & \cellcolor{lightblue}(5190.8) & \cellcolor{lightblue}(2136.8) & \cellcolor{lightblue}(7814.2) & \cellcolor{lightblue}[13.3\%] & \cellcolor{lightblue}[18.1\%] \\
       \multirow{2}{*}{TIES-Merging}
       & 89.3 & 80.0 & 40.1 & 44.9 & 45.1 & 26.7 & 54.3 & 58.2 \\
       & \cellcolor{lightblue}(712.5) & \cellcolor{lightblue}(1837.0) & \cellcolor{lightblue}(948.6) & \cellcolor{lightblue}(3140.4) & \cellcolor{lightblue}(1231.7) & \cellcolor{lightblue}(6002.2) & \cellcolor{lightblue}[38.3\%] & \cellcolor{lightblue}[49.5\%] \\
       \hline
       \rowcolor{lightgray}
       \multicolumn{9}{c}{\textit{Activation-informed Merging (ACM)}} \\
       \multirow{2}{*}{ACM-Average}
       & 90.7 & 81.8 & 40.8 & 45.6 & 47.4 & 26.7 & 55.5 & 60.0 \\
       & \cellcolor{lightblue}(742.2) & \cellcolor{lightblue}(1337.3) & \cellcolor{lightblue}(816.7) & \cellcolor{lightblue}(2218.0) & \cellcolor{lightblue}(937.9) & \cellcolor{lightblue}(4273.0) & \cellcolor{lightblue}[57.5\%] & \cellcolor{lightblue}[60.9\%] \\
       \multirow{2}{*}{ACM-TA}
       & 92.6 & 69.4 & 30.9 & 34.8 & 32.0 & 26.7 & 47.7 & 49.8 \\
       & \cellcolor{lightblue}(1545.6) & \cellcolor{lightblue}(4374.3) & \cellcolor{lightblue}(4903.2) & \cellcolor{lightblue}(7501.5) & \cellcolor{lightblue}(3892.2) & \cellcolor{lightblue}(7965.4) & \cellcolor{lightblue}[$-$24.2\%] & \cellcolor{lightblue}[$-$39.2\%] \\
       \multirow{2}{*}{ACM-TIES}
       & 91.2 & 81.6 & 39.0 & 46.5 & 47.5 & 26.7 & 55.4 & 60.2 \\
       & \cellcolor{lightblue}(640.6) & \cellcolor{lightblue}(1273.7) & \cellcolor{lightblue}(742.2) & \cellcolor{lightblue}(2145.8) & \cellcolor{lightblue}(899.8) & \cellcolor{lightblue}(5069.6) & \cellcolor{lightblue}[55.7\%] & \cellcolor{lightblue}[62.9\%] \\
       \hline
       \rowcolor{lightgray}
       \multicolumn{9}{c}{\textit{Single-Objective Evolutionary Merging}} \\
       \multirow{2}{*}{Evo-L2S (Accuracy)}
       & \textbf{95.1} & \textbf{89.0} & 43.8 & 51.7 & 45.7 & 33.3 & 59.8 & \textbf{61.7 }\\
       & \cellcolor{lightblue}(945.7) & \cellcolor{lightblue}(2663.1) & \cellcolor{lightblue}(3027.3) & \cellcolor{lightblue}(5406.2) & \cellcolor{lightblue}(2307.6) & \cellcolor{lightblue}(7570.7) & \cellcolor{lightblue}[9.8\%] & \cellcolor{lightblue}[12.1\%] \\
       \multirow{2}{*}{Evo-L2S (Length)}
       & 89.1 & 74.0 & 34.6 & 38.7 & 45.2 & 13.3 & 49.1 & 56.6 \\
       & \cellcolor{lightblue}(367.5) & \cellcolor{lightblue}(695.0) & \cellcolor{lightblue}(832.7) & \cellcolor{lightblue}(1120.2) & \cellcolor{lightblue}(699.5) & \cellcolor{lightblue}(1246.2) & \cellcolor{lightblue}[79.6\%] & \cellcolor{lightblue}[75.6\%] \\
       \hline
       \rowcolor{lightgray}
       \multicolumn{9}{c}{\textit{Multi-Objective Evolutionary Merging (Entropy Sampling, Pareto-optimal)}} \\
       \multirow{2}{*}{Evo-L2S-01}
       & 94.3 & 86.4 & 43.8 & 49.8 & 45.8 & 33.3 & 58.9 & 61.1 \\
       & \cellcolor{lightblue}(858.5) & \cellcolor{lightblue}(2429.8) & \cellcolor{lightblue}(2742.5) & \cellcolor{lightblue}(5318.2) & \cellcolor{lightblue}(2129.9) & \cellcolor{lightblue}(7797.2) & \cellcolor{lightblue}[12.5\%] & \cellcolor{lightblue}[17.6\%] \\
       \multirow{2}{*}{Evo-L2S-02}
       & 94.8 & 85.4 & 43.4 & 49.9 & 46.4 & 26.7 & 57.8 & 61.4 \\
       & \cellcolor{lightblue}(763.6) & \cellcolor{lightblue}(2272.6) & \cellcolor{lightblue}(2429.9) & \cellcolor{lightblue}(5051.1) & \cellcolor{lightblue}(1965.9) & \cellcolor{lightblue}(7589.9) & \cellcolor{lightblue}[17.4\%] & \cellcolor{lightblue}[23.6\%] \\
       \multirow{2}{*}{Evo-L2S-03}
       & 93.3 & 83.8 & 40.4 & 45.9 & 46.4 & 26.7 & 56.1 & 60.3 \\
       & \cellcolor{lightblue}(593.4) & \cellcolor{lightblue}(1958.2) & \cellcolor{lightblue}(1957.3) & \cellcolor{lightblue}(4871.0) & \cellcolor{lightblue}(1700.6) & \cellcolor{lightblue}(7301.5) & \cellcolor{lightblue}[24.4\%] & \cellcolor{lightblue}[32.4\%] \\
       \multirow{2}{*}{Evo-L2S-04}
       & 94.2 & 83.0 & 40.4 & 47.0 & \textbf{47.7} & 16.7 & 54.8 & 61.2 \\
       & \cellcolor{lightblue}(273.9) & \cellcolor{lightblue}(1008.8) & \cellcolor{lightblue}(700.8) & \cellcolor{lightblue}(1885.5) & \cellcolor{lightblue}(840.0) & \cellcolor{lightblue}(4090.9) & \cellcolor{lightblue}[63.8\%] & \cellcolor{lightblue}[69.3\%] \\
       \multirow{2}{*}{Evo-L2S-05}
       & 91.5 & 77.2 & 37.9 & 37.3 & 46.1 & 20.0 & 51.7 & 57.9 \\
       & \cellcolor{lightblue}(245.2) & \cellcolor{lightblue}(680.2) & \cellcolor{lightblue}(815.2) & \cellcolor{lightblue}(1053.8) & \cellcolor{lightblue}(696.0) & \cellcolor{lightblue}(1299.9) & \cellcolor{lightblue}[80.3\%] & \cellcolor{lightblue}[77.1\%] \\
    \bottomrule[0.8pt]
    \end{tabular}
    }
    \caption{Results of all model merging methods at the 14B scale.}
    \label{tab:14B_res}
\end{table}

\section{Qualitative Reasoning Example}
\label{app:qualitative}

Beyond aggregate output length, we qualitatively inspect how Evo-L2S changes the reasoning traces produced by the model. \Cref{tab:qualitative_trace} shows a representative MATH500 example for which both \texttt{DeepSeek-R1-Distill-Qwen-7B} and \texttt{Evo-L2S-02} reach the correct answer. The reasoning baseline repeatedly revisits and verifies an already established solution, whereas Evo-L2S follows a more direct trajectory with substantially less redundant deliberation. This example illustrates how the reduction in generated tokens can correspond to more concise reasoning rather than simply truncating the solution.

\begin{table}[t]
\centering
\begin{tabular}{p{0.10\textwidth}p{0.4\textwidth}p{0.4\textwidth}}
\toprule
\textbf{Problem} &
\texttt{\textbf{DeepSeek-R1-Distill-Qwen-7B}} &
\texttt{\textbf{Evo-L2S-02}} \\
\midrule
MATH500 IDX=16 &
Okay, so I have this math problem here: Compute \(1 - 2 + 3 - 4 + 5 - \dots + 99 - 100\). Hmm, let me try to figure this out step by step. First, I notice that the series alternates between adding and subtracting consecutive numbers. It starts with 1, then subtracts 2, adds 3, subtracts 4, and so on, all the way up to subtracting 100. So, the pattern is \(+1, -2, +3, -4, \dots, +99, -100\). I wonder if there's a way to group these terms to make it easier to compute. Maybe if I pair them up? Let me try that. If I group the first two terms: \(1 - 2\), that equals \(-1\). Then the next two terms: \(3 - 4\), which is also \(-1\). Hmm, interesting. Let me check the next pair: \(5 - 6 = -1\). It seems like every pair of consecutive numbers adds up to \(-1\). So, if each pair is \(-1\), how many such pairs are there in the series? Well, the series goes from 1 to 100, so that's 100 terms in total. If I pair them two at a time, that would be \(100 / 2 = 50\) pairs. Therefore, if each pair is \(-1\), the total sum should be \(50 \times (-1) = -50\). Wait, let me make sure I didn't make a mistake here. So, starting from 1, subtracting 2, adding 3, subtracting 4, etc., up to 100. So, each pair is indeed two numbers, one positive and one negative, with the positive being odd and the negative being even. Let me write out the first few pairs to confirm:1 - 2 = -13 - 4 = -15 - 6 = -17 - 8 = -1...99 - 100 = -1. Yes, each pair is -1, and there are 50 such pairs because 100 divided by 2 is 50. So, 50 times -1 is -50. Is there another way to think about this? [...] the final answer is: \boxed{-50} &
To solve the problem \(1 - 2 + 3 - 4 + 5 - \dots + 99 - 100\), we can break it down into manageable steps. 1. **Understand the Pattern:**   The sequence alternates between positive and negative terms. Specifically, the terms are \(1, -2, 3, -4, 5, -6, \dots, 99, -100\). 2. **Group the Terms:**   We can group the terms in pairs: \((1 - 2), (3 - 4), (5 - 6), \dots, (99 - 100)\). Each pair sums to \(-1\). 3. **Count the Pairs:**   Since we are pairing every two terms, and there are 100 terms in total, there will be \(100 / 2 = 50\) pairs. 4. **Sum the Pairs:**   Since each pair sums to \(-1\), the total sum will be \(50 \times (-1) = -50\). Let's implement this in Python to verify our reasoning.```python\# Calculate the sum of the series 1 - 2 + 3 - 4 + 5 - ... + 99 - 100 total\_sum = 0\# Iterate through the series for i in range(1, 101):    if i \% 2 == 0:  \# if i is even, subtract it total\_sum -= i    else:  \# if i is odd, add it        total\_sum += i print(total\_sum)``````output-50``` The sum of the series \(1 - 2 + 3 - 4 + 5 - \dots + 99 - 100\) is \(\boxed{-50}\).  \\
\bottomrule
\end{tabular}
\caption{Illustrative reasoning-trace comparison. Both models produce the correct answer. \texttt{DeepSeek-R1-Distill-Qwen-7B}'s output is shortened by 3317 characters for readability. For both generations we remove special characters (e.g., newline).}
\label{tab:qualitative_trace}
\end{table}

\end{document}